\PassOptionsToPackage{hidelinks}{hyperref}
\documentclass[letterpaper, 10 pt, conference]{ieeeconf}  

\IEEEoverridecommandlockouts                              

\usepackage{bm}
\usepackage{graphicx} 
\usepackage{amsmath}
\usepackage{amsfonts}
\usepackage{amssymb}
\usepackage{bm}
\usepackage{algorithm}
\usepackage{algpseudocode}
\usepackage{xcolor}
\usepackage{algorithm}
\usepackage{algpseudocode}
\usepackage{subcaption}
\usepackage{tabularx}
\usepackage{amsmath}
\usepackage{amssymb}
\usepackage{mathtools}
\usepackage{mathrsfs}
\usepackage{bbm}
\usepackage{bm}
\usepackage{siunitx}

\usepackage{booktabs}
\usepackage{multirow}
\usepackage{array}
\usepackage{makecell}
\usepackage{wrapfig}

\usepackage{graphicx}
\usepackage{adjustbox}
\usepackage{svg}

\usepackage[table,dvipsnames]{xcolor}

\usepackage{tikz}
\usetikzlibrary{fit,calc,positioning}
\usepackage{pgfplots}
\pgfplotsset{compat=1.16}

\usepackage{pbox}
\usepackage{xspace}
\usepackage{multicol}
\usepackage{microtype}
\usepackage{bbding}
\usepackage{listings}

\usepackage[hidelinks]{hyperref}

\newcommand{\mypar}[1]{\par\noindent\textbf{#1}\ }

\usepackage{caption}

\newcommand{\OurMethod}{\textsc{Mosaic}}

\title{\LARGE \bf
\OurMethod{}: Mode- and Support-Aware Inference-Time Control of Robot Diffusion Models
}

\author{Sabrina Bodmer$^{1}$, Rahel Rickenbach$^{1}$, René Zurbrügg$^{1}$, Melanie N. Zeilinger$^{\dagger1}$, Colin Jones$^{\dagger2}$
\thanks{This research has been supported by the Swiss National Science Foundation under the NCCR Automation (grant 51NF40\_225155) and an ETH AI Center doctoral fellowship awarded to René Zurbrügg.}
\thanks{$^{1}$Department of Mechanical and Process Engineering, ETH Z\"urich, Switzerland.
        {\tt\small sabodmer@ethz.ch}}%
\thanks{$^{2}$Automatic Control Laboratory, EPFL, Switzerland.}%
\thanks{$^{\dagger}$ Denotes equal supervision.}%
}

\begin{document}

\maketitle
\thispagestyle{empty}
\pagestyle{empty}

\begin{abstract}
Generative policies such as diffusion models provide expressive data-driven priors for robot behavior, yet the situations a robot faces at execution time rarely match its training data and impose constraints and objectives the policy never learned. Existing inference-time guidance methods either rely on gradients that can push samples away from the learned prior, or on sampling procedures that can be inefficient, collapse modes, or average them into out-of-distribution predictions.

We introduce \OurMethod, a mode- and support-aware framework for gradient-free inference-time control of robotic diffusion models. 
\OurMethod{} treats the initial noise and all stochastic reverse-step perturbations as a joint input to the pretrained denoising process and adapts a distribution over this input using importance weights computed from fully denoised outputs. 
Physical feasibility and consistency with the learned support are balanced using adaptive penalties, while mode-wise importance-weighted updates preserve modes and avoid destructive averaging.

The same formulation can jointly steer multiple independently trained diffusion models through shared coupling costs, enabling coordinated behavior without requiring a joint model. We evaluate \OurMethod{} on navigation, dexterous grasping, motion planning with unseen geometry, and collaborative box manipulation with independently trained robot priors. Across these settings, \OurMethod{} adapts and composes frozen diffusion priors under new deployment constraints without retraining.\\
Videos and additional results available at: \href{https://mosaic-guide.github.io/}{mosaic-guide.github.io}
\end{abstract}

\section{INTRODUCTION}
Diffusion models have emerged as powerful generative priors for robotic decision-making, capturing complex and multimodal distributions over grasps, trajectories, and action sequences from offline data. Their ability to represent multiple valid behaviors is particularly valuable in robotics, where a task may admit different configurations, homotopy classes, or coordinated behaviors. 
However, deployment inevitably introduces requirements that were absent during training, such as new obstacles, embodiment-specific constraints, or interactions with independently trained agents.
Reusing a frozen diffusion prior therefore requires adapting its predictions to new deployment conditions without discarding the behavioral structure learned from data. \\
\begin{figure}[t]
    \centering
    \includegraphics[width=1.0\linewidth]{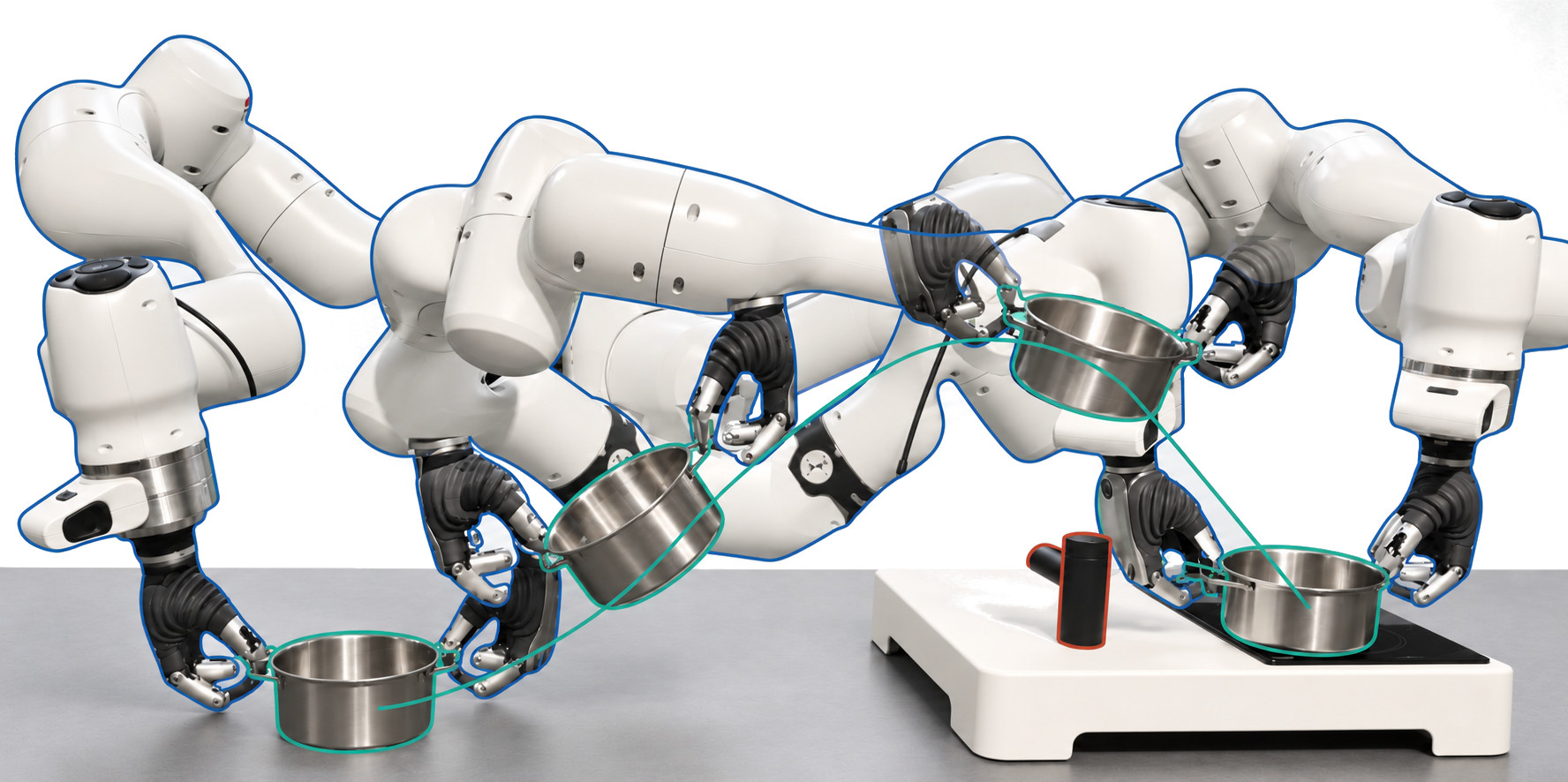}
\caption{
\textbf{Inference-time composition.}
\OurMethod{} couples independent diffusion priors through a shared objective, enabling coordinated behavior without joint training.}
    \label{fig:teaser} \vspace{-4mm}
\end{figure}
The inference-time adaptation problem has several coupled difficulties. First, many relevant objectives, such as collision checking, inverse kinematics, and grasp stability, are non-differentiable, expensive to evaluate, or only meaningful for a fully denoised prediction. 
Second, an output may satisfy a physical objective while leaving the support of the learned prior, which can produce physically admissible yet implausible or semantically invalid behavior. 
Third, the learned prior itself may be multimodal, and global sampling updates can average incompatible solution families into low-density or infeasible predictions. 
For example, averaging paths that pass on opposite sides of an obstacle can produce a path through the obstacle. Addressing feasibility alone is therefore insufficient; the inference procedure must also preserve support and multimodality.

Existing guidance methods address parts of this problem. However,  gradient-based approaches require differentiable objectives and can push samples away from the support of the learned distribution. And while 
sampling-based methods avoid this requirement and admit arbitrary black-box costs, sequential particle methods can suffer from weight degeneracy and loss of particle diversity under repeated reweighting and resampling~\cite{del2006sequential}. In diffusion steering, this difficulty is compounded when candidates are ranked on estimates of the eventual clean output at highly noisy intermediate states; recent work has observed that such value estimates can be inaccurate early in the reverse process~\cite{jain2026diffusion}.
Delaying guidance to later denoising steps makes terminal estimates more reliable, but leaves fewer reverse transitions through which earlier decisions can be revised.

We introduce \OurMethod{}, a mode- and support-aware framework for gradient-free inference-time control of frozen robot diffusion models. 
\OurMethod{} treats the complete reverse process as a mapping from its stochastic inputs to a clean prediction. We expose the initial noise and all reverse-step perturbations as a joint control input and iteratively adapt a distribution over these inputs using model predictive path integral (MPPI)-style importance-weighted updates.
The approach evaluates complete behaviors rather than intermediate denoising states, allowing deployment constraints to be assessed where they are most meaningful. 
To preserve the structure learned by the diffusion prior, \OurMethod{} balances physical feasibility with consistency to the learned support and maintains separate proposals for distinct solution modes, avoiding destructive averaging across incompatible behaviors. 
The same formulation also allows independently trained diffusion priors to be coordinated through shared inference-time costs, enabling multi-agent composition without training a joint model.

In summary, our contributions are:
\begin{enumerate}
\item \emph{Complete reverse-process control.} 
We expose the initial diffusion sample and all reverse-step perturbations as a joint control input, enabling arbitrary deployment objectives to be evaluated on fully denoised rollouts.

\item \emph{Mode- and support-aware proposal adaptation.}
We approximate the resulting constrained distribution using mode-specific importance-weighted proposals and adapt physical and support constraint weights online.

\item \emph{Inference-time composition of independent priors.} We show that the same framework can coordinate independently trained diffusion models through shared coupling objectives, enabling multi-agent behavior without retraining a joint model.
\end{enumerate}

\section{Related Work}
\label{sec:related_work}
\mypar{Gradient-based diffusion guidance.}
Gradient-based methods steer the reverse process using derivatives of a reward, constraint, or auxiliary objective with respect to the diffusion state~\cite{bansal2023universalguidancediffusionmodels,ma25constraintawarediff,umionair2025,bayesfk2026}. For example, CoDiG incorporates barrier-function gradients for collision avoidance~\cite{ma25constraintawarediff}, while UMI-on-Air uses gradients of an embodiment-specific tracking objective to bias manipulation trajectories toward dynamically executable solutions~\cite{umionair2025}. 
Related optimization-based approaches enforce deployment-time constraints
by optimizing reverse-process variables or solving constrained problems
around denoised predictions~\cite{bodmer2026grounding,giaretta2026direct}. These approaches often require differentiable objectives, surrogates, or optimization during denoising, and may move samples away from regions supported by the learned diffusion prior.

\mypar{Sampling-based diffusion guidance.}
Sampling-based methods avoid differentiating through the deployment objective by evaluating and comparing candidate predictions directly, making them applicable to non-differentiable constraints and objectives. One class maintains a population of candidates and progressively concentrates computation on promising samples through selection, mutation, reweighting, or resampling~\cite{yang2024diffusionES,wu2024practicalasymptoticallyexactconditional,singhal2025generalframeworkinferencetimescaling,skreta2025feynmankaccorrectorsdiffusionannealing,wang2026twistSMC}. Feynman--Kac steering provides a sequential formulation of this idea, where particles are weighted and resampled throughout the reverse process using intermediate potentials, often constructed from denoised estimates of the eventual clean prediction~\cite{singhal2025generalframeworkinferencetimescaling,skreta2025feynmankaccorrectorsdiffusionannealing}. Such sequential updates can work well when intermediate estimates are reliable, but repeated reweighting and resampling can quickly concentrate the population on only a few candidates. If promising alternatives receive low weight early in denoising, they may disappear before their final behavior can be evaluated, leading to particle starvation and mode collapse.

A second class uses evaluated samples to iteratively adapt the distribution from which subsequent candidates are drawn~\cite{kim2026ZeNO,DNO2024,park2026grace,xiang2026NDPP_Grasp}. 
Such proposal adaptation is closely related to MPPI, cross-entropy optimization, and adaptive importance sampling~\cite{williams2017mppi, Boer2005, cappe2008adaptive}, which update parametric sampling distributions from weighted candidate populations.
GRACE, for example, uses MPPI-style importance weighting to update Gaussian proposals during the reverse process~\cite{park2026grace}, while ZeNO applies a path-integral update to the source noise of a deterministic diffusion sampler~\cite{kim2026ZeNO}. By concentrating the proposal around promising regions, these methods can progressively improve candidate solutions without maintaining particle lineages throughout the reverse process. However, when a single proposal is updated from samples belonging to multiple distinct solution modes, its global statistics can combine incompatible behaviors~\cite{honda2024stein}. Such averaging can move subsequent samples toward low-density or infeasible regions and lead to loss of multimodality. In contrast, \OurMethod{} treats the source noise and all stochastic reverse-step perturbations jointly as the proposal variable, evaluates them only after complete denoising, and maintains separate proposals for distinct output modes.

\mypar{Composition of diffusion models.}
Coordinating multiple diffusion models typically requires either learning their interaction during training or combining independently generated predictions at inference time~\cite{karim2025dagdiff,guo2025unidiffgrasp}. DAGDiff learns a joint diffusion model over dual-arm grasp pairs and uses classifier guidance to steer the coupled prediction~\cite{karim2025dagdiff}. UniDiffGrasp instead generates candidates for the two arms independently and subsequently selects compatible grasp pairs~\cite{guo2025unidiffgrasp}. Joint models require task-specific training, while post-hoc selection can require many samples when compatible pairs are rare. 
More generally, Feynman--Kac Correctors show how pretrained diffusion models can be combined through product-distribution constructions at inference time~\cite{skreta2025feynmankaccorrectorsdiffusionannealing}. \OurMethod{} instead introduces coordination through an arbitrary terminal coupling cost,
which need not correspond to a separately trained score or differentiable energy model.

\section{Background}
\label{sec:background}

\subsection{Diffusion Models and DDIM Sampling}
\label{sec:diffusion_background}

Let $x_0\sim p_{\mathrm{data}}$ denote a clean robot trajectory, grasp, or action sequence, and $c$ the conditioning context, e.g., the robot's environment.
The forward process of a diffusion model produces a noisy state
\begin{equation}
    x_t = \sqrt{\bar\alpha_t}\,x_0 + \sqrt{1-\bar\alpha_t}\,\varepsilon, \qquad \varepsilon\sim\mathcal{N}(0,I),
    \label{eq:forward}
\end{equation}
and the denoising network $\hat\varepsilon_\theta(x_t,t;c)$ predicts the added noise, yielding the clean estimate at step $t$,
\begin{equation}
    \hat x_0(x_t,t;c)=\frac{x_t -\sqrt{1-\bar\alpha_t}\,\hat\varepsilon_\theta(x_t,t;c)}{\sqrt{\bar\alpha_t}}.
    \label{eq:tweedie}
\end{equation}
A stochastic DDIM sampler transitions from $x_t$ to $x_{t-1}$ as
\begin{equation}
    x_{t-1}=\mu_\theta(x_t,t;c)+\sigma_t\omega_t,\qquad\omega_t\sim\mathcal{N}(0,I),
    \label{eq:ddim_reverse}
\end{equation}
with its deterministic part given as
\begin{align}
    \mu_\theta(x_t,t;c)={}&\sqrt{\bar\alpha_{t-1}}\,\hat x_0(x_t,t;c)\nonumber\\
    &+\sqrt{1-\bar\alpha_{t-1}-\sigma_t^2}\,\hat\varepsilon_\theta(x_t,t;c).
    \label{eq:ddim_mean}
\end{align}
Standard sampling draws a fresh $\omega_t$ at every step.
Thus, a complete reverse-process is determined by the initial noise $x_T$ together with the stochastic perturbations $\omega_T,\ldots,\omega_1$. 

\subsection{Model Predictive Path-Integral Control}
\label{sec:mppi}

Model predictive path integral control (MPPI)~\cite{williams2017mppi}
optimizes system inputs using sampled rollouts rather than gradients through the dynamics.
MPPI repeatedly samples around a nominal input, evaluates the resulting rollouts, and shifts the sampling distribution toward lower-cost inputs.
We index these proposal updates by $\ell$.

At iteration $\ell$, MPPI maintains a Gaussian proposal
\begin{equation}
    q_\ell(v)=\mathcal{N}(v;\bar v_\ell,\Sigma),
\end{equation}
centered at the current nominal input $\bar v_\ell$.
A population is sampled as
\begin{equation}
    v_\ell^{(i)}=\bar v_\ell+\epsilon_\ell^{(i)},\quad\epsilon_\ell^{(i)}\sim\mathcal{N}(0,\Sigma),\quad i=1,\ldots,S,
\end{equation}
and each sample is evaluated through a rollout with cost $C^{(i)}$.
Lower-cost samples thereby receive larger normalized weights,
\begin{equation}
    w_\ell^{(i)}
    =
    \frac{
        \exp\!\left(-C^{(i)}/\lambda\right)
    }{
        \sum_{j=1}^{S}
        \exp\!\left(-C^{(j)}/\lambda\right)
    },
\end{equation}
and the nominal input is shifted toward the weighted population,
\begin{equation}
    \bar v_{\ell+1}
    =
    \bar v_\ell
    +
    \eta_v
    \sum_{i=1}^{S}
    w_\ell^{(i)}
    \left(
        v_\ell^{(i)}-\bar v_\ell
    \right).
    \label{eq:mppi_background}
\end{equation}

Repeated sampling and reweighting therefore moves the proposal toward inputs that produce lower-cost rollouts.

\section{Problem Formulation}
\label{sec:problem}
We consider a frozen diffusion model that, for a context $c$, generates robot outputs $x_0$ by iterated denoising. Each run of the reverse process returns one such output, and repeated sampling induces the model distribution $p_\theta(x_0\mid c)$ whose samples must satisfy objectives and constraints introduced only at deployment.
Specifically, we seek samples with low task cost $J_{\rm task}$ while
satisfying physical $J_{\rm phys}$ and support $J_{\rm sup}$ requirements
\begin{align}
    \min_{x_0 \sim p_\theta(\cdot\mid c)} \quad
        & J_{\rm task}(x_0;c)
    \label{eq:constrained_problem1}\\
    \mathrm{s.t.}\quad
        & J_{\rm phys}(x_0;c)
        \leq \epsilon_{\rm phys}, \nonumber\\
        & J_{\rm sup}(x_0;c)
        \leq \epsilon_{\rm sup}. \nonumber
\end{align}
The support cost $J_{\rm sup}$ can be any scalar measure of how atypical
$x_0$ is under the learned prior, e.g., a negative log-likelihood under a
density model fitted to training samples in output or feature space.
Accordingly, $\epsilon_{\rm sup}$ can be calibrated from the support scores of reference samples. 

These criteria may be non-differentiable and are often meaningful only on
the final prediction. Our goal is therefore to adapt the inference
distribution without retraining the model or differentiating through the
denoising process or deployment costs.

\section{Method}
\label{sec:method}
\begin{figure*}[t]
    \centering
    \includegraphics[width=0.95\textwidth, trim={0cm 0.4cm 0cm 0.4cm}, clip]{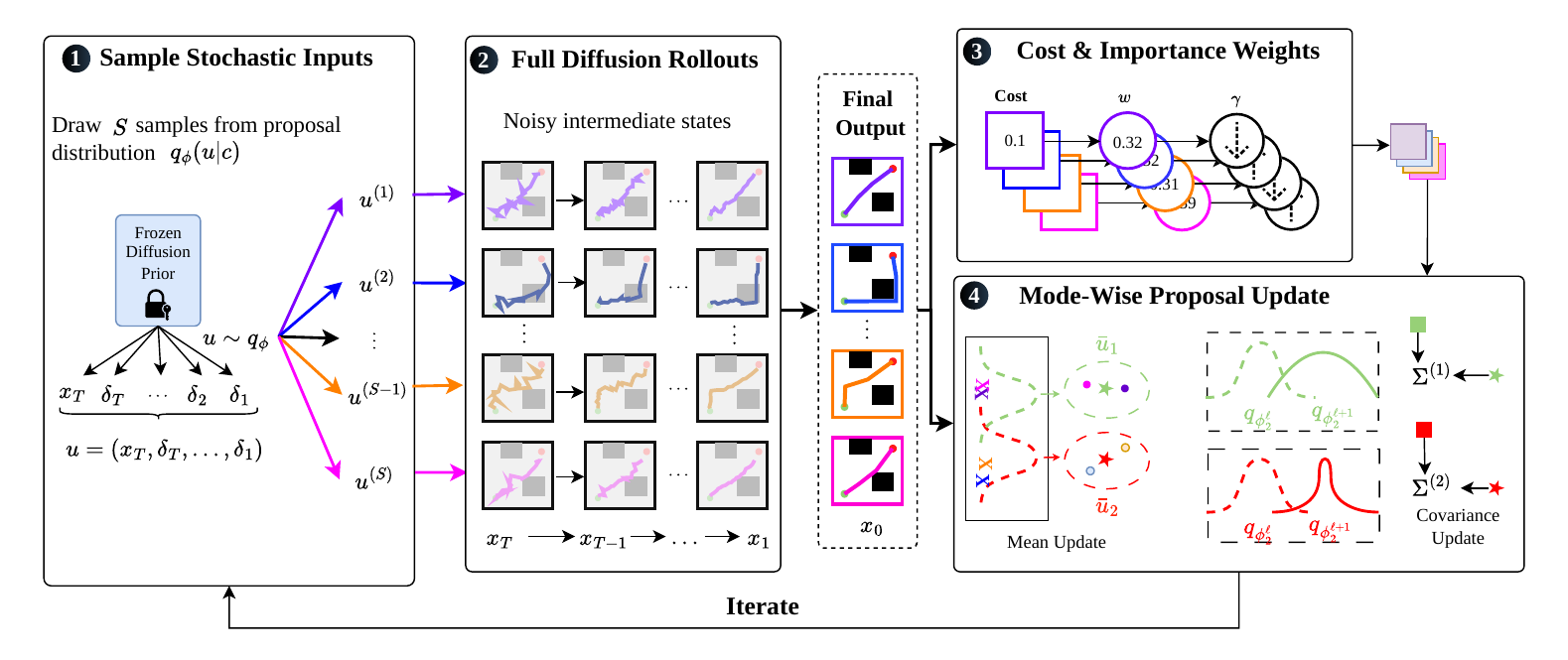}
    \caption{\textbf{\OurMethod{} evaluates complete denoising rollouts and updates each solution mode separately.} A reverse-process input consists of the initial diffusion noise $x_T$ and all reverse-step perturbations $\delta_T, \dots, \delta_1$. The frozen model maps this input to a clean prediction $x_0$, on which arbitrary task, physical, support, and coupling costs can be evaluated. The clean predictions are grouped into modes, and a separate importance-weighted update is performed for each mode.}
    \label{fig:overview}
    \vspace{-6mm}
\end{figure*}

\OurMethod{} approaches this problem by adapting the stochastic inputs of the complete reverse process. 
We first collect the initial noise and all reverse-step perturbations into a joint reverse input $u$,  making the denoising process deterministic. 
We then formulate a desired deployment-time distribution over this input,
balancing task performance, physical and support constraints, and deviation from the pretrained prior (Sec.~\ref{sec:constrained_reverse_input}).
Since this target cannot in general be sampled from directly, we approximate it using importance-weighted proposals. To avoid averaging incompatible solution families, the proposal is maintained
separately for modes identified from the fully denoised outputs (Sec.~\ref{sec:mode_aware_updates}). 
Finally, the same construction extends to several independent diffusion priors by evaluating their complete rollouts under a shared coupling cost (Sec.~\ref{sec:composition}). The complete procedure is illustrated in Fig.~\ref{fig:overview}.

\subsection{Complete Reverse Process as a Black-Box Rollout}
\label{sec:controlled_rollout}
Following the setup above, we collect the stochastic quantities in Eq.~\eqref{eq:ddim_reverse} into a single \emph{reverse input}
\begin{equation}
    u=(x_T,\bm\delta),
    \qquad \bm\delta=(\delta_T,\ldots,\delta_1),
    \label{eq:reverse_input}
\end{equation}
where $\delta_t=\omega_t$.
For fixed $u$ and context $c$, the reverse process is deterministic and defines the rollout map 
\begin{equation}
    x_0=\mathcal{G}_\theta(u;c),
    \label{eq:rollout_map}
\end{equation}
from the complete reverse input $u$ to the final clean prediction $x_0$. This reformulation converts inference-time guidance into a sampling-based control problem over the complete stochastic input of the reverse process. Rather than modifying individual denoising transitions, we optimize a distribution over $u= (x_T,\delta_T,\ldots,\delta_1)$, with each candidate evaluated only after propagation through the complete denoising rollout.

Given independent sampling of each stochastic transition vector $\delta_t$, the reference distribution over reverse inputs is modeled as
\begin{equation}
    p_{\mathrm{ref}}(u)
    =\mathcal{N}(x_T;0,I)
     \prod_{t=1}^{T}\mathcal{N}(\delta_t;0,I).
    \label{eq:reference_input}
\end{equation} 
Sampling $u\sim p_{\rm ref}$ and applying $\mathcal G_\theta$ therefore
reproduces the unguided DDIM output distribution $p_\theta(x_0\mid c)$,
making $p_{\rm ref}$ the natural reference distribution over reverse inputs.
Starting from this reference, \OurMethod{} seeks a deployment-time distribution that additionally satisfies the objectives in Sec.~\ref{sec:problem}.
We formulate this distribution next.

\subsection{Constrained Reverse-Input Target}
\label{sec:constrained_reverse_input}
Given our task and support requirements $J_{\mathrm{task}}$, $J_{\mathrm{phys}}$, and $J_{\mathrm{sup}}$, as well as the map from a complete reverse input $u$ to its clean output $x_0$, we relax the sample-level constraints to expectation constraints over the reverse-input distribution. The resulting problem is
\begin{align}\label{eq:constrained_problem}
    \min_{q \ll p_{\mathrm{ref}}}\quad
        &\mathbb{E}_q[J_{\mathrm{task}}(\mathcal{G}_\theta(u;c))] + \lambda D_{\mathrm{KL}}(q||p_\mathrm{ref})\\
    \mathrm{s.t.}\quad
        &\mathbb{E}_q[J_{\mathrm{phys}}(\mathcal{G}_\theta(u;c))]
        \leq \varepsilon_{\mathrm{phys}},  \nonumber\\
        &\mathbb{E}_q[J_{\mathrm{sup}}(\mathcal{G}_\theta(u;c))]
        \leq \varepsilon_{\mathrm{sup}}. \nonumber
\end{align}
Its Lagrangian follows as 
\begin{align}
    \mathcal{L}(q, \gamma) = &\mathbb{E}_q[J_{\mathrm{task}}] + \lambda D_{\mathrm{KL}}(q||p_{\mathrm{ref}}) \\
    &+\mathbb{E}_q[\sum_{j \in \{\mathrm{phys}, \mathrm{sup}\}} \gamma_j (J_j - \varepsilon_j)]. \nonumber
\end{align}
For fixed multipliers $\gamma$, the optimization over $q$ has the standard
KL-regularized variational form. By the Gibbs variational identity underlying
information-theoretic path-integral control~\cite{williams2017mppi}, its
minimizer is the exponentially tilted reference distribution
\begin{align}
    p^*_{\gamma}(u|c) = \frac{1}{Z_\gamma}p_{\mathrm{ref}}(u) \exp\left[-\frac{C_\gamma(\mathcal{G}_\theta(u;c);c)}{\lambda}\right],
\end{align}
where 
\begin{align}
    C_\gamma = J_{\mathrm{task}} + \sum_{j \in \{\mathrm{phys}, \mathrm{sup}\}} \gamma_j J_j,
\end{align}
and $G_\theta$ is the rollout map.
Thus, low-cost reverse inputs are more likely under the target distribution, while $p_{\rm ref}$ anchors the adapted distribution to the stochastic inputs of the frozen prior.
The resulting target can be approximated using adaptive importance sampling~\cite{cappe2008adaptive}, applied here to the complete stochastic input of the frozen reverse diffusion process. 
Following projected dual ascent~\cite{boyd2004convex} and related
primal--dual diffusion formulations~\cite{hadou2026constraineddiffusionmodelsprimaldual, tomasi2026primaldualguideddecodingconstrained}, the ideal multiplier update is
\begin{align}\label{eq:ideal_dual_update}
    \gamma_j^{\ell+1} = \left[\gamma_j^\ell + \eta_{\gamma,j}(\mathbb{E}_{p_{\gamma^\ell}^*}[J_j] - \varepsilon_j) \right]_+.
\end{align}
A constraint that exceeds its budget therefore receives a larger penalty
in the next target distribution, while an inactive constraint can relax.
Thus, the target distribution $p_\gamma^*$ is defined by $\gamma^\ell$ at the current iteration, while the expected constraint violations determine the target at the next iteration. \OurMethod{} estimates both from the same finite population of complete rollouts.

\subsection{Mode-Aware Proposal Adaptation} 
\label{sec:mode_aware_updates}
At iteration \(\ell\), MOSAIC approximates the current target \(p_{\gamma^\ell}^*\) using samples drawn from an adaptive proposal. 
A single-proposal update in Eq.~\eqref{eq:mppi_background} can be
problematic when the target is multimodal:
samples from distinct solution families may all receive high weight,
while their weighted mean corresponds to neither family. We therefore
replace the single proposal by a mixture of $M$ mode-specific components,
\begin{equation}\label{eq:mixture_proposal}
    q_{\phi^\ell}(u\mid c)=\sum_{m=1}^M\pi_m^\ell q_{\phi_m^\ell}(u\mid c),
\end{equation}
where $\pi_m^\ell$ denotes the probability of sampling from mode $m$. Each component \(q_{\phi_m^\ell}\) is a factorized Gaussian over the complete reverse input $u$, with mode-specific mean \(\bar u_m^\ell\) and covariance \(\Sigma_m^\ell\). 
At \(\ell=0\), we sample from \(p_{\rm ref}\) and cluster the corresponding fully denoised outputs.
These assignments initialize the mode centers, mixture
masses, and mode-conditioned reverse-input means, with covariance initialized
from the reference proposal. Centers may then remain fixed or be re-estimated from the current population at selected iterations. 
Subsequent iterations sample from the updated mixture $q_{\phi^\ell}$.

\mypar{Importance Weighting.} 
At each iteration, samples are drawn from the current proposal
$q_{\phi^\ell}$, but we want their weighted population to represent the target $p_{\gamma^\ell}^*$.
Each sample $i$ is therefore weighted by the ratio between its target density and its proposal density. For $u_\ell^{(i)}\sim q_{\phi^\ell}$, we propagate each candidate through the complete reverse process, and evaluate \(C_{\gamma^\ell}\) on its fully denoised output. 
Using the target definition above, and accounting for the full mixture density in Eq.~\eqref{eq:mixture_proposal}, the importance weights are
\begin{align}
    \tilde w_\ell^{(i)} &= \exp\left[-\frac{C_{\gamma^\ell}^{(i)}}{\lambda}\right]\frac{p_{\mathrm{ref}}(u_\ell^{(i)})}{q_{\phi^\ell}(u_\ell^{(i)} | c)}, \label{eq:importance_weight} \\
    w_\ell^{(i)} &= \frac{\tilde w_\ell^{(i)}}{\sum_{j=1}^S \tilde w_\ell^{(j)}}. \label{eq:normalized_weight}
\end{align}
Under the usual support and integrability assumptions, these self-normalized
importance weights consistently estimate expectations under
$p_{\gamma^\ell}^*$~\cite{Geweke1989BayesianII}.

\mypar{Constraint adaptation.}
The above importance weights $w_\ell^{(i)}$  estimate expectations under the target \(p_{\gamma^\ell}^*\), including the constraint expectations required by Eq.~\eqref{eq:ideal_dual_update}:
\begin{align}\label{eq:constraint_cost_estimate}
    \hat J_{j, \ell} = \sum_{i=1}^S w_\ell^{(i)}J_j(x_{0|\ell}^{(i)};c).
\end{align}
Replacing the expectation in Eq.~\eqref{eq:ideal_dual_update} by its importance-sampling estimate gives
\begin{align}\label{eq:dual_update}
    \gamma_j^{\ell+1} = \left[\gamma_j^\ell + \eta_{\gamma,j}(\hat J_{j,\ell} - \varepsilon_j)\right]_+.
\end{align}

\mypar{Mode-wise mean adaptation.} 
We embed each fully denoised output in a task-dependent representation
\(h_\ell^{(i)}=e_\phi(x_{0|\ell}^{(i)},c)\)
and assign it to the current mode centers. If online re-clustering is enabled, the centers are first re-estimated from the current population and aligned with the previous modes. The resulting assignments are represented by probabilities $r_{im}^\ell\in[0,1]$, with $\sum_m r_{im}^\ell=1$. A hard assignment is recovered when each $r_{im}^\ell$ is either zero or one. The global weight $w_\ell^{(i)}$ determines how desirable sample $i$ is, while $r_{im}^\ell$ determines how strongly it belongs to mode $m$.

\noindent The estimated importance mass associated with mode $m$ and the normalized weight within that mode are then given as
\vspace{-1mm}
\begin{equation}
    \hat\rho_m^\ell=\sum_{i=1}^{S}w_\ell^{(i)} r_{im}^\ell,
    \qquad
    w_{i\mid m}^\ell=\frac{r_{im}^\ell\tilde w_\ell^{(i)}}
    {\sum_{s=1}^{S}r_{sm}^\ell\tilde w_\ell^{(s)}}.
    \label{eq:mode_weights}
\end{equation}

The conditional weights $w_{i|m}^{\ell}$ sum to one separately for every mode. Consequently, the reverse-input mean of each mode can be updated without averaging it with samples assigned to another mode:
\vspace{-1mm}
\begin{align} \label{eq:mode_update}
    \bar u_m^{\ell+1}
    &=\bar u_m^\ell+\eta_u\sum_{i=1}^{S}w_{i\mid m}^\ell
       \bigl(u_\ell^{(i)}-\bar u_m^\ell\bigr).
\end{align}

\mypar{Exploration adaptation.}
We additionally update the block-wise covariance of each reverse step. For reverse block \(t\), let \(b_{t,m}^\ell\) denote the median intermediate cost among samples assigned to mode \(m\), and let \(\bar C_{t,m}^{\ell+1}\) denote the same cost evaluated on the intermediate prediction of the updated component mean. We set
\begin{align}\label{eq:mode_cov_update}
    \Sigma_{t|\ell+1}^{(m)}&=\exp(\frac{\bar C_{t,m}^{\ell+1}-b_{t,m}^{\ell}}{\lambda_\Sigma})\sigma_\delta^2 I. 
\end{align}
Thus, exploration contracts around promising component centers and expands around poorly performing ones. Note that intermediate predictions are used only to adapt exploration variance; importance weighting, constraint adaptation, and final selection remain based on fully denoised outputs.

Together, the importance-weighted mean update in Eq.~\eqref{eq:mode_update} and the cost-adaptive exploration rule in Eq.~\eqref{eq:mode_cov_update} define the proposal for mode \(m\) at the next iteration,
\begin{align}
    q_{\phi_m^{\ell+1}}(u\mid c)={}&\mathcal{N}(x_T;\bar x_{T|\ell+1}^{(m)}, \Sigma_{T|\ell+1}^{x,(m)}) \nonumber\\
    &\times \prod_{t=1}^{T}\mathcal{N}(\delta_t;\bar\delta_{t|\ell+1}^{(m)},\Sigma_{t|\ell+1}^{\delta,(m)}).
    \label{eq:updated_mode_proposal}
\end{align}

\mypar{Sample allocation.}
A total of $S$ samples are drawn from $M$ modes. 
We update the probability of sampling from mode $m$ at iteration $\ell+1$ according to 
\begin{equation}
    \pi_m^{\ell+1}
    \propto \max\!\left(
    (1-\eta_\pi)\pi_m^\ell+\eta_\pi\hat\rho_m^\ell,\pi_{\min}\right),
    \label{eq:mixture_mass_update}
\end{equation}
followed by normalization across modes. 
The update allocates more of the next finite sample budget to modes with larger estimated target mass, while \(\pi_{\min}\) prevents temporary low weight from permanently eliminating a mode.

After \(L\) proposal iterations, each component mean is rolled out independently, and we return the valid mode with minimum task cost; if none satisfies the budgets, the query is marked infeasible.
The full method is shown in Algorithm~\ref{alg:mosaic}.

\begin{figure*}[t]
    \vspace{5mm}
    \centering
    \includegraphics[width=0.95\linewidth,
        trim={0 0.2cm 0 0.5cm},
        clip]{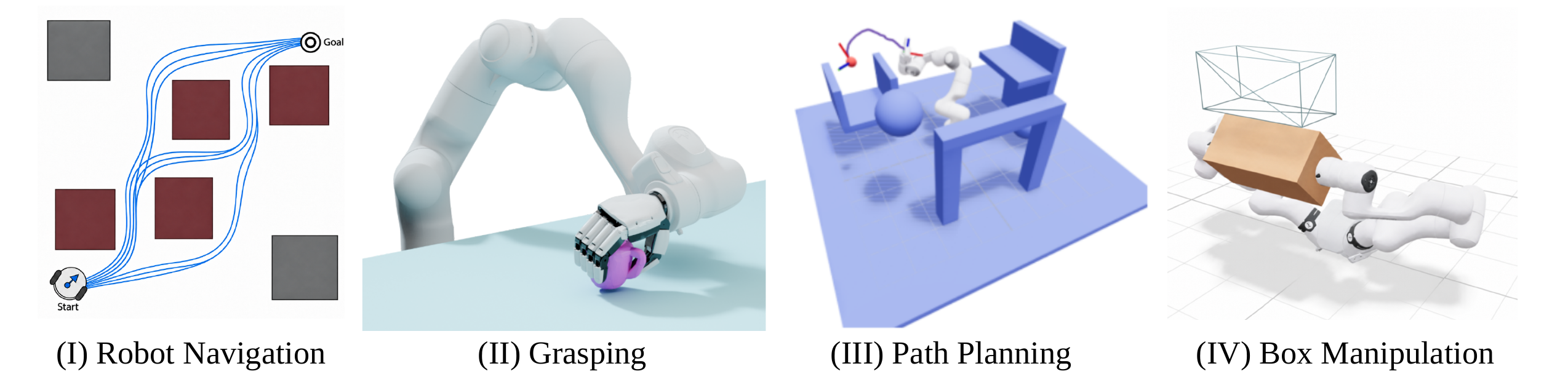}
    \caption{\textbf{Overview of the evaluated tasks.}
    From left to right:
    (I) \textit{Robot Navigation}, where a mobile robot must find feasible paths through a 2D environment;
    (II) \textit{Grasping}, where a robotic manipulator must generate collision-free grasps in clutter;
    (III) \textit{Path Planning}, where a robot arm must find a collision-free trajectory from a start to a goal configuration;
    and (IV) \textit{Collaborative Box Manipulation}, where two arms must jointly move a box to a target pose without dropping or crushing it.}
    \label{fig:tasks} \vspace{-5mm}
\end{figure*}

\begin{algorithm}[b]
\caption{\OurMethod{}}
\label{alg:mosaic}
\begin{footnotesize}
\begin{algorithmic}[1]
\Require Frozen rollout $\mathcal{G}_\theta$, context $c$, samples $S$, iterations $L$, budgets $\varepsilon_j$, modes $M$
\State Initialize $q_{\phi^0}=p_{\mathrm{ref}}$ and $\gamma_j^0=0$
\For{$\ell=0,\ldots,L-1$}
    \State Sample $u_\ell^{(i)}\sim q_{\phi^\ell}$ for $i=1,\ldots,S$
    \State Roll out $x_{0|\ell}^{(i)}=\mathcal{G}_\theta(u_\ell^{(i)};c)$
    \State Evaluate terminal costs $C_{\gamma^{\ell}}^{(i)}$ and compute importance weights $\tilde w_\ell^{(i)}, w_\ell^{(i)}$ using
    Eq.~\eqref{eq:importance_weight}, .~\eqref{eq:normalized_weight}
    \State Estimate $\widehat J_{j, \ell}$ and update $\gamma_j^{\ell+1}$ using~\eqref{eq:constraint_cost_estimate},~\eqref{eq:dual_update}
    \State Embed $x_{0|\ell}^{(i)}$; optionally re-cluster and align modes; compute assignments $r_{im}^\ell$
    \State Compute $w_{i|m}^\ell$ and $\hat\rho_m^\ell$ 
    using Eq.~\eqref{eq:mode_weights}
    \State Update means and covariance $\bar u_{m}^{\ell+1}$,$\Sigma^{(m)}_{t|\ell+1}$ using Eq.~\eqref{eq:mode_update},~\eqref{eq:mode_cov_update}
    \State Update mixture masses $\pi_m^{\ell+1}$ using Eq.~\eqref{eq:mixture_mass_update}
    \State Form $q_{\phi^{\ell+1}} = \sum_m \pi_m^{\ell+1}q_{\phi_m^{\ell+1}}$
\EndFor
\State Roll out all final mode means and return the best valid prediction
\end{algorithmic}
\end{footnotesize}
\end{algorithm}

\subsection{Compositional Coupling}
\label{sec:composition}
Because the importance weights depend only on costs evaluated after complete rollouts, the same construction can jointly steer several frozen diffusion priors.  Consider a collaborative setting, such as dual-arm systems, with reverse inputs $u^L$ and $u^R$, and define $u^{\mathrm{joint}} = [(u^L)^\top \; (u^R)^\top]^\top$.
Each model is rolled out with its own frozen denoiser, $x_0^a=\mathcal{G}_{\theta_a}(u^a;c^a)$ for $a\in\{L,R\}$. The two clean outputs are evaluated together using
\begin{equation}
    C_{\gamma}^{\mathrm{joint}}
    =C_{\gamma^L}^L(x_0^L)+C_{\gamma^R}^R(x_0^R)
     +C_{\mathrm{couple}}(x_0^L,x_0^R),
    \label{eq:joint_cost}
\end{equation}
where $C_{\gamma^L}^L$ and $C_{\gamma^R}^R$ measure individual quality and $C_{\mathrm{couple}}$ measures compatibility between the two predictions. For bimanual manipulation, the coupling term can encode relative contacts, object-level constraints, or collisions between the two hands. 

Assuming that the stochastic inputs of the two unguided samplers are drawn
independently, their joint reference distribution factorizes as
\begin{align} \label{eq:joint_reference}
    p_{\mathrm{ref}}^{\mathrm{joint}}(u^L, u^R) &= p_{\mathrm{ref}}^L(u^L)p_{\mathrm{ref}}^R(u^R).
\end{align}
The coupling cost then induces the non-factorized target
\begin{align}
    p^*_{\gamma,\mathrm{joint}}(u^L,u^R) &\propto p_{\mathrm{ref}}^{\mathrm{joint}}\exp \left(-\frac{C_\gamma^{\mathrm{joint}}(u^L, u^R)}{\lambda}\right).
\end{align}

The pair receives one joint importance weight, which updates both $u^L$ and $u^R$. This steers the models toward compatible combinations while each hand remains anchored to its pretrained prior, eliminating the need for a bimanual diffusion model or modification of the individual denoisers.
\vspace{-1mm}

\section{Experiments}
\label{sec:experiments}

\begin{table*}[t]
\vspace{3mm}

\centering

\caption{\textbf{Results across all tasks.}
SR denotes success rate, UNI unique solutions, COLL maze-wall collisions,
$H$ mode entropy, and SPL success weighted by path length.
Values are mean $\pm$ std over three seeds (navigation: eight mazes with $16$ trajectories each; planning and box manipulation: $50$ trials per seed), except grasping, which reports binomial standard error over $280$ grasps. Sampling-based methods use matched initial populations and DDIM steps. \OurMethod{} preserves navigation diversity while support reduces wall collisions. Its success advantage over the strongest baseline grows from grasping to planning and is largest for box manipulation, which couples independently trained priors at inference time.
}
\label{tab:main_results}

\setlength{\tabcolsep}{3pt}
\renewcommand{\arraystretch}{1.1}
\resizebox{0.95\textwidth}{!}{%
\begin{tabular}{l|ccc|cc||cc|cc|cc}
\toprule
\multirow{3}{*}{\textbf{Method}}
 & \multicolumn{3}{c|}{\multirow{2}{*}{\textbf{2D Navigation}}}
 & \multicolumn{2}{c|}{\multirow{2}{*}{\textbf{Grasping}}}
 & \multicolumn{4}{c|}{\textbf{Trajectory Planning}}
 & \multicolumn{2}{c}{\multirow{2}{*}{\textbf{Box Manipulation}}} \\
\cmidrule(lr){7-10}
 & \multicolumn{3}{c|}{}
 & \multicolumn{2}{c|}{}
 & \multicolumn{2}{c|}{w/o tool}
 & \multicolumn{2}{c|}{w/ tool}
 & \multicolumn{2}{c}{} \\
\cmidrule(lr){2-4}
\cmidrule(lr){5-6}
\cmidrule(lr){7-8}
\cmidrule(lr){9-10}
\cmidrule(lr){11-12} & $\mathrm{SR}\,\uparrow_{\%}$
 & $\mathrm{UNI}\,\uparrow_{\%}$
 & $\mathrm{COLL}\,\downarrow_{\%}$
 & $\mathrm{SR}\,\uparrow_{\%}$
 & $H\,\uparrow_{-}$
 & $\mathrm{SR}\,\uparrow_{\%}$
 & $\mathrm{SPL}\,\uparrow_{\%}$
 & $\mathrm{SR}\,\uparrow_{\%}$
 & $\mathrm{SPL}\,\uparrow_{\%}$
 & $\mathrm{SR}\,\uparrow_{\%}$
 & $\mathrm{SPL}\,\uparrow_{\%}$ \\
\midrule
\rowcolor{gray!5}
DDIM~\cite{DDIMsong2023denoising}
 & 13.8 {\tiny $\pm$ 3.5} & 99.2 {\tiny $\pm$ 0.8} & 38.8 {\tiny $\pm$ 3.2}
 & 19.6 {\tiny $\pm$ 2.4} & 0.75 {\tiny $\pm$ 0.11}
 & 46.7 {\tiny $\pm$ 7.0} & 25.0 {\tiny $\pm$ 5.0} & 44.0 {\tiny $\pm$ 6.0} & 24.6 {\tiny $\pm$ 4.6}
 & 0.0 {\tiny $\pm$ 0.0} & 0.0 {\tiny $\pm$ 0.0} \\

Gradient Guidance
 & \ 4.4 {\tiny $\pm$ 3.3} & 99.5 {\tiny $\pm$ 0.9} & 50.3 {\tiny $\pm$ 3.5}
 & 27.5 {\tiny $\pm$ 2.7} & 0.37 {\tiny $\pm$ 0.08}
 & 52.7 {\tiny $\pm$ 8.1} & 26.8 {\tiny $\pm$ 4.0} & 52.7 {\tiny $\pm$ 10.1} & 26.0 {\tiny $\pm$ 4.5}
 & 0.0 {\tiny $\pm$ 0.0} & 0.0 {\tiny $\pm$ 0.0} \\
\midrule

\rowcolor{gray!5}
FK Steering~\cite{singhal2025generalframeworkinferencetimescaling}
 & \textbf{90.9 {\tiny $\pm$ 9.1}} & 55.7 {\tiny $\pm$ 10.2} & \underline{\ 7.6 {\tiny $\pm$ 1.6}}
 & \underline{51.1 {\tiny $\pm$ 3.0}} & 1.00 {\tiny $\pm$ 0.09}
 & 51.3 {\tiny $\pm$ 5.0} & 28.0 {\tiny $\pm$ 4.3} & 50.0 {\tiny $\pm$ 6.0} & 28.0 {\tiny $\pm$ 2.5}
 & 24.0 {\tiny $\pm$ 10.4} & 17.0 {\tiny $\pm$ 7.0} \\

Path-Space MPPI
 & 45.8 {\tiny $\pm$ 3.5} & 98.4 {\tiny $\pm$ 0.0} & 27.6 {\tiny $\pm$ 3.7}
 & 21.1 {\tiny $\pm$ 2.4} & 0.42 {\tiny $\pm$ 0.09}
 & 58.7 {\tiny $\pm$ 8.1} & 31.5 {\tiny $\pm$ 5.4} & 52.7 {\tiny $\pm$ 6.4} & 28.4 {\tiny $\pm$ 2.1}
 & 29.3 {\tiny $\pm$ 4.2} & 24.7 {\tiny $\pm$ 4.1} \\
\rowcolor{gray!5}

NDPP~\cite{xiang2026NDPP_Grasp}
 & 41.7 {\tiny $\pm$ 3.9} & 95.8 {\tiny $\pm$ 1.2} & 30.7 {\tiny $\pm$ 3.2}
 & 32.9 {\tiny $\pm$ 2.8} & 0.83 {\tiny $\pm$ 0.09}
 & 76.7 {\tiny $\pm$ 4.6} & 41.8 {\tiny $\pm$ 2.2} & 71.3 {\tiny $\pm$ 12.1} & 40.3 {\tiny $\pm$ 6.3}
 & 42.7 {\tiny $\pm$ 4.2} & 32.4 {\tiny $\pm$ 3.9} \\
\midrule

\rowcolor{blue!8}
\OurMethod{} (Ours)
 & \underline{80.7 {\tiny $\pm$ 0.9}} & \underline{99.5 {\tiny $\pm$ 0.5}} & \textbf{\ 3.9 {\tiny $\pm$ 0.0}}
 & \textbf{53.2 {\tiny $\pm$ 3.0}} & \underline{1.03 {\tiny $\pm$ 0.09}}
 & \underline{86.7 {\tiny $\pm$ 3.1}} & \underline{47.9 {\tiny $\pm$ 1.7}} & \underline{79.3 {\tiny $\pm$ 3.1}} & \underline{46.3 {\tiny $\pm$ 2.0}}
 & \textbf{78.7 {\tiny $\pm$ 6.1}} & \textbf{67.6 {\tiny $\pm$ 5.1}} \\

\rowcolor{gray!5}
\quad w/o support cost
 & 67.2 {\tiny $\pm$ 0.8} & \textbf{99.7 {\tiny $\pm$ 0.5}} & 23.2 {\tiny $\pm$ 1.2}
 & 50.4 {\tiny $\pm$ 3.0} & \textbf{1.04 {\tiny $\pm$ 0.07}}
 & \textbf{87.3 {\tiny $\pm$ 2.3}} & \textbf{48.2 {\tiny $\pm$ 1.6}} & \textbf{81.3 {\tiny $\pm$ 1.2}} & \textbf{47.7 {\tiny $\pm$ 0.8}}
 & \underline{78.0 {\tiny $\pm$ 5.3}} & \underline{67.2 {\tiny $\pm$ 4.7}} \\
\bottomrule
\end{tabular}%
}
\vspace{-5mm}
\end{table*}
We evaluate four capabilities of \OurMethod{}: preserving multimodal solutions, adapting to unseen geometry, generating executable grasps, 
and composing independent priors.

\mypar{Tasks.}
In \textit{1)} 2D navigation, inference-time obstacles constrain a multimodal trajectory prior with multiple homotopy classes using the model and code from~\cite{luo2024potential}. 
\textit{2)} Dexterous grasping adapts a frozen Franka--XHand grasp prior  from~\cite{zurbrugg2026dexevolve} using kinematic and collision constraints. 
\textit{3)} Trajectory planning and \textit{4)} collaborative box manipulation use the diffusion model from~\cite{carvalho2025motion}; the former introduces unseen collision geometry and tool attachment, while the latter couples two independently trained arm priors on the real dual-arm system. All pretrained generative models remain frozen.

\mypar{Baselines and protocol.}
We compare against unguided DDIM~\cite{DDIMsong2023denoising}, gradient guidance, Path-Space MPPI, FK steering~\cite{singhal2025generalframeworkinferencetimescaling}, and NDPP~\cite{xiang2026NDPP_Grasp}. Path-Space MPPI uses a single Gaussian proposal over the complete reverse input; FK steering
sequentially reweights and resamples using intermediate clean-output
estimates; and NDPP updates the current reverse correction from finite-horizon
rollouts evaluated through Tweedie-predicted clean outputs.

Within each task, all methods use the same frozen model, number of DDIM steps, and initial sample count. Since the methods expose different inference-time choices, we tune these separately and report the results of the best-performing configuration for each task, including the NDPP horizon, FK resampling schedule, mean versus best-sample selection for MPPI and \OurMethod{}, and optional online re-clustering for \OurMethod{}. Task-level DDIM horizons and initial populations are selected during preliminary tuning and then held fixed across methods.

Success is task-specific: in 2D navigation, a trajectory is successful only if none of its $48$ waypoints collides with an obstacle. 
Grasping success requires arm IK feasibility, no hand penetration of the table plane, and stability under simulated forces.
Trajectory-planning considers solutions successful if the trajectory is collision-free and the final end-effector position error is below $3$\,cm. Box manipulation requires final box position and orientation errors below $2$\,cm and $5^\circ$, respectively, with the distance between the two hands varying by at most $3$\,cm throughout the motion. We also report success weighted by path length (SPL), using the straight-line start--goal distance as the reference length; failed trials contribute zero.
Each method is evaluated on identical task instances. Navigation uses $8$ mazes with $16$ trajectories each over $3$ seeds ($384$ trajectories total); trajectory planning and box manipulation use $50$ trials per seed over $3$ seeds ($150$ trials per condition); and grasping uses a total of $280$ grasps over $7$ objects. We report standard deviation across seeds for navigation, trajectory planning, and box manipulation, and binomial standard error for grasping.

\mypar{Implementation details.}
For navigation, grasping, trajectory planning, and box manipulation, we use $(8,4,10,10)$ DDIM steps and $(48,48,64,64)$ initial samples, respectively. The larger horizons and populations are used for the more involved planning and multi-arm tasks.  Support models are fitted offline, with budgets calibrated from the unguided population at $\ell=0$. Navigation modes use a path signature, while trajectory and box modes use subsampled end-effector path signatures; mode centers may remain fixed or be re-clustered online depending on the selected configuration.

\begin{figure}
    \centering
    \includegraphics[width=0.9\linewidth, trim={0cm 0cm 0cm 0cm}, clip]{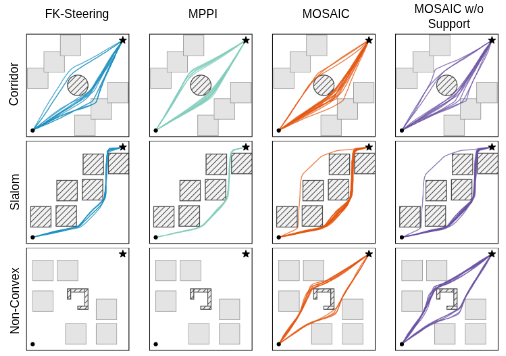}
    \caption{
    \textbf{2D navigation example.} Rows show corridor, slalom, and non-convex environments; columns compare FK
    steering, Path-Space MPPI, \OurMethod{}, and \OurMethod{} without support.
    Shaded obstacles are maze walls encoded by the prior, while striped obstacles
    are added only at inference time.
    The black dot and star denote start and goal. FK steering and Path-Space MPPI can lose distinct modes, whereas \OurMethod{} retains multiple feasible solution families; support regularization further reduces collisions with maze walls.
    }
    \label{fig:maze}
    \vspace{-5mm}
\end{figure}

\vspace{-2mm}
\subsection{Results} 
\label{sec:results}
Table~\ref{tab:main_results} summarizes the quantitative results.

\mypar{Mode preservation.}
The 2D navigation task isolates mode starvation and support drift under
inference-time steering. FK steering achieves the highest success
($90.9\pm9.1\%$), but retains only $55.7\pm10.2\%$ unique solutions,
showing substantial particle starvation across solution modes. In contrast,
\OurMethod{} reaches $80.7\pm0.9\%$ success while retaining
$99.5\pm0.5\%$ unique solutions. Path-Space MPPI also preserves diversity
($98.4\%$), but reaches only $45.8\pm3.5\%$ success, illustrating the
difficulty of adapting a single global proposal across distinct modes.
Fig.~\ref{fig:maze} shows these behaviors qualitatively.

Support regularization is particularly important here because the maze walls
are encoded only by the learned prior and do not appear in the deployment
cost. With support, \OurMethod{} reduces wall collisions from
$23.2\pm1.2\%$ to $3.9\pm0.0\%$ while increasing success from
$67.2\pm0.8\%$ to $80.7\pm0.9\%$, with essentially unchanged diversity.
This shows that the support constraint can preserve task structure learned by
the prior that is invisible to the deployment objective.

\mypar{Dexterous grasping.}
\OurMethod{} obtains the highest grasp success at $53.2\%$, compared with $51.1\pm3.0\%$ for FK steering and $32.9\%$ for NDPP, while retaining high mode entropy. Removing support reduces success to $50.4\%$ with essentially unchanged entropy, suggesting that support regularization improves grasp feasibility without sacrificing diversity.

\mypar{Deployment-time geometry.}
For trajectory planning, \OurMethod{} reaches  $86.7\pm3.1\%$ success without the tool and $79.3\pm3.1\%$ with unseen tool geometry, compared with $76.7\pm4.6\%$ and $71.3\pm12.1\%$ for the strongest baseline, NDPP. 
\OurMethod{} also achieves higher success-weighted path efficiency in both
settings ($47.9\%/46.3\%$ versus $41.8\%/40.3\%$).

\mypar{Inference-time composition.}
On collaborative box manipulation, \OurMethod{} reaches
$78.7\pm6.1\%$ success, compared with $42.7\pm4.2\%$ for the strongest
baseline, NDPP. It also achieves the highest success-weighted path efficiency
($67.6\pm5.1\%$ versus $32.4\pm3.9\%$), despite using independently trained
single-arm priors with no jointly trained bimanual model.

\begin{figure}[t]
    \vspace{3mm}

    \centering
    \includegraphics[width=0.95\linewidth]{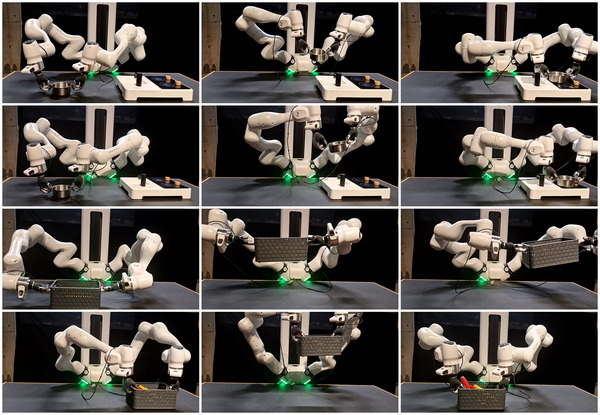}
  \caption{
  \textbf{Real-world collaborative manipulation.}
    Each row shows one \OurMethod{} rollout from grasp establishment (left) through transport to the target pose (right). The two arms use independently trained frozen diffusion priors coupled only at inference time through Eq.~\eqref{eq:joint_cost}.
    \label{fig:rollouts}}
    \vspace{-3mm}
\end{figure}

\mypar{Hardware deployment.}
We deploy the bimanual composition task on a dual-arm Franka setup equipped with BrainCo hands (Fig.~\ref{fig:teaser}), using one independently trained single-arm diffusion model per arm. The hand configurations are fixed rather than generated: we use a pinch grasp for a small pan and flat-hand or regular grasps for a larger box. These setups induce different required wrist and arm configurations during transport. In each case, the object is lifted and moved to a new target while the two arms maintain feasible relative distances and orientations throughout the motion.  Representative hardware rollouts are shown in Fig.~\ref{fig:rollouts}.


\mypar{Effect of rollout-level evaluation.}
A clear trend emerges across the manipulation tasks: the advantage of
\OurMethod{} is small for grasping, increases for trajectory planning, and is
largest for collaborative manipulation. In the latter tasks, success depends
increasingly on coupled properties of the complete behavior, such as
swept-volume collisions, whole-trajectory feasibility, or coordination between
two arms. Intermediate clean-output or Tweedie estimates used by sequential
methods such as FK steering and NDPP may become less informative proxies for
these rollout-level criteria, whereas \OurMethod{} evaluates them directly on
complete rollouts. This may explain part of the observed trend;
the remaining gap to Path-Space MPPI, which also uses complete rollouts,
suggests benefit from mode-wise proposal adaptation, consistent
with the mode-starvation observed in navigation.

\section{Limitations}
\label{sec:limitations}
While \OurMethod{} performs well across the evaluated tasks, some limitations remain. Evaluating complete rollouts introduces an inference-time overhead of approximately 2$\times$ relative to comparable sampling-based baselines, which may become significant in high-rate closed-loop settings. Calibrating the support model requires prior training data, or alternatively unguided model samples at the cost of a less faithful density estimate. The task-specific mode embedding and fixed number of modes motivate learned embeddings and adaptive mode creation. The reverse-input perspective could also be extended beyond diffusion models, e.g., to flow matching, which we leave to future work.

\vspace{-1mm}
\section{Conclusion}
\label{sec:conclusion}
We presented \OurMethod{}, a gradient-free framework that adapts the complete reverse input of frozen robot diffusion models. Costs are evaluated on fully denoised outputs, physical and support constraints are balanced through adaptive multipliers, and mode-specific proposals preserve distinct solution families. 
The same formulation couples independently trained priors through a shared
terminal objective without training a joint model.

Across navigation, grasping, trajectory planning, and bimanual manipulation,
\OurMethod{} preserves multimodal solutions while adapting frozen priors to
deployment-time constraints. It reaches $86.7\%$ trajectory-planning success,
retains $79.3\%$ with unseen tool geometry, and achieves $78.7\%$ success in
collaborative manipulation compared with $42.7\%$ for the strongest baseline.
Reducing the cost of complete rollouts and learning more general mode
representations are natural directions for future work.

\section*{\small ACKNOWLEDGEMENTS}

{\small
The author(s) used generative AI to assist with drafting and editing the manuscript, cleaning up figures and coding. All content was reviewed by the authors, who take full responsibility for it. \\
We thank Alexandre Didier for helpful discussions during the early stages of this project. The authors also thank Agile Robots for generously providing the Mobile FR3 Duo robot used for this work.
\par}
\bibliographystyle{ieeetr}
\bibliography{main}

@misc{xiang2026NDPP_Grasp,
      title={NDPP-Grasp: Non-Differentiable Physical Plausibility Constraint-Guided Task-Oriented Dexterous Grasp Generation}, 
      author={Qiuchi Xiang and Haoxuan Qu and Hossein Rahmani and Jun Liu},
      year={2026},
      eprint={2606.02432},
      archivePrefix={arXiv},
      primaryClass={cs.RO},
      url={https://arxiv.org/abs/2606.02432}, 
}

@misc{park2026grace,
      title={GRACE: Gradient-Free Robot Action Generation via Combined Diffusion-MPPI Posterior Mean Estimation}, 
      author={Leesai Park and Jiho HOng and Sanghyun Kim},
      year={2026},
      eprint={2607.21661},
      archivePrefix={arXiv},
      primaryClass={cs.RO},
      url={https://arxiv.org/abs/2607.21661}, 
}

@misc{DNO2024,
      title={Inference-Time Alignment of Diffusion Models with Direct Noise Optimization}, 
      author={Zhiwei Tang and Jiangweizhi Peng and Jiasheng Tang and Mingyi Hong and Fan Wang and Tsung-Hui Chang},
      year={2024},
      eprint={2405.18881},
      archivePrefix={arXiv},
      primaryClass={cs.LG},
      url={https://arxiv.org/abs/2405.18881}, 
}

@misc{kim2026ZeNO,
      title={Gradient-Free Noise Optimization for Reward Alignment in Generative Models}, 
      author={Jeongsol Kim and Hongeun Kim and Jian Wang and Jong Chul Ye},
      year={2026},
      eprint={2605.11347},
      archivePrefix={arXiv},
      primaryClass={cs.LG},
      url={https://arxiv.org/abs/2605.11347}, 
}

@misc{yang2024diffusionES,
      title={Diffusion-ES: Gradient-free Planning with Diffusion for Autonomous Driving and Zero-Shot Instruction Following}, 
      author={Brian Yang and Huangyuan Su and Nikolaos Gkanatsios and Tsung-Wei Ke and Ayush Jain and Jeff Schneider and Katerina Fragkiadaki},
      year={2024},
      eprint={2402.06559},
      archivePrefix={arXiv},
      primaryClass={cs.LG},
      url={https://arxiv.org/abs/2402.06559}, 
}

@misc{wang2026twistSMC,
      title={Inference-Time Alignment of Diffusion Models via Trust-Region Iterative Twisted Sequential Monte Carlo}, 
      author={Weixin Wang and Yu Yang and Wei Deng and Pan Xu},
      year={2026},
      eprint={2605.25123},
      archivePrefix={arXiv},
      primaryClass={cs.LG},
      url={https://arxiv.org/abs/2605.25123}, 
}

@misc{karim2025dagdiff,
      title={DAGDiff: Guiding Dual-Arm Grasp Diffusion to Stable and Collision-Free Grasps}, 
      author={Md Faizal Karim and Vignesh Vembar and Keshab Patra and Gaurav Singh and K Madhava Krishna},
      year={2025},
      eprint={2509.21145},
      archivePrefix={arXiv},
      primaryClass={cs.RO},
      url={https://arxiv.org/abs/2509.21145}, 
}

@misc{guo2025unidiffgrasp ,
      title={UniDiffGrasp: A Unified Framework Integrating VLM Reasoning and VLM-Guided Part Diffusion for Open-Vocabulary Constrained Grasping with Dual Arms}, 
      author={Xueyang Guo and Hongwei Hu and Chengye Song and Jiale Chen and Zilin Zhao and Yu Fu and Bowen Guan and Zhenze Liu},
      year={2025},
      eprint={2505.06832},
      archivePrefix={arXiv},
      primaryClass={cs.RO},
      url={https://arxiv.org/abs/2505.06832}, 
}

@misc{tomasi2026primaldualguideddecodingconstrained,
      title={Primal-Dual Guided Decoding for Constrained Discrete Diffusion}, 
      author={Federico Tomasi and Dmitrii Moor and Alice Wang and Mounia Lalmas},
      year={2026},
      eprint={2605.09749},
      archivePrefix={arXiv},
      primaryClass={cs.AI},
      url={https://arxiv.org/abs/2605.09749}, 
}

@misc{hadou2026constraineddiffusionmodelsprimaldual,
      title={Constrained Diffusion Models with Primal-Dual Inference}, 
      author={Samar Hadou and Yigit Berkay Uslu and Alejandro Ribeiro},
      year={2026},
      eprint={2606.17192},
      archivePrefix={arXiv},
      primaryClass={cs.LG},
      url={https://arxiv.org/abs/2606.17192}, 
}

@article{DDIMsong2023denoising,
  title={{Denoising Diffusion Implicit Models}},
  author={Song, Jiaming and Meng, Chenlin and Ermon, Stefano},
  journal = {arXiv preprint arXiv:2010.02502},
  year = {2020},
}

@article{bodmer2026grounding,
  title={Grounding Generative Policies in Physics: Optimization-Guided Diffusion for Robot Control},
  author={Bodmer, Sabrina and Zurbr{\"u}gg, Ren{\'e} and Portela, Tifanny and Ma, Hao and Didier, Alexandre and Hutter, Marco and Jones, Colin and Zeilinger, Melanie},
  journal={arXiv preprint arXiv:2606.24208},
  year={2026}
}

@misc{bayesfk2026,
      title={BayesFP: Posterior Estimation for Flow-Based Policies via Feynman-Kac Sampling}, 
      author={Sreevardhan Sirigiri and Weiming Zhi and Fabio Ramos},
      year={2026},
      eprint={2606.21014},
      archivePrefix={arXiv},
      primaryClass={cs.RO},
      url={https://arxiv.org/abs/2606.21014}, 
}

@misc{wu2024practicalasymptoticallyexactconditional,
      title={Practical and Asymptotically Exact Conditional Sampling in Diffusion Models}, 
      author={Luhuan Wu and Brian L. Trippe and Christian A. Naesseth and David M. Blei and John P. Cunningham},
      year={2024},
      eprint={2306.17775},
      archivePrefix={arXiv},
      primaryClass={stat.ML},
      url={https://arxiv.org/abs/2306.17775}, 
}

@misc{singhal2025generalframeworkinferencetimescaling,
      title={A General Framework for Inference-time Scaling and Steering of Diffusion Models}, 
      author={Raghav Singhal and Zachary Horvitz and Ryan Teehan and Mengye Ren and Zhou Yu and Kathleen McKeown and Rajesh Ranganath},
      year={2025},
      eprint={2501.06848},
      archivePrefix={arXiv},
      primaryClass={cs.LG},
      url={https://arxiv.org/abs/2501.06848}, 
}

@misc{skreta2025feynmankaccorrectorsdiffusionannealing,
      title={Feynman-Kac Correctors in Diffusion: Annealing, Guidance, and Product of Experts}, 
      author={Marta Skreta and Tara Akhound-Sadegh and Viktor Ohanesian and Roberto Bondesan and Alán Aspuru-Guzik and Arnaud Doucet and Rob Brekelmans and Alexander Tong and Kirill Neklyudov},
      year={2025},
      eprint={2503.02819},
      archivePrefix={arXiv},
      primaryClass={cs.LG},
      url={https://arxiv.org/abs/2503.02819}, 
}

@InProceedings{ma25constraintawarediff,
  title = 	 {Constraint-{A}ware {D}iffusion {G}uidance for {R}obotics: {R}eal-{T}ime {O}bstacle {A}voidance for {A}utonomous {R}acing},
  author =       {Ma, Hao and Bodmer, Sabrina and Carron, Andrea and Zeilinger, Melanie and Muehlebach, Michael},
  booktitle = 	 {CoRL},
  pages = 	 {},
  year = 	 {2025},
}

@ariticle{umionair2025,
      title={{UMI}-on-{A}ir: {E}mbodiment-{A}ware {G}uidance for {E}mbodiment-{A}gnostic {V}isuomotor {P}olicies}, 
      author={Harsh Gupta and Xiaofeng Guo and Huy Ha and Chuer Pan and Muqing Cao and Dongjae Lee and Sebastian Scherer and Shuran Song and Guanya Shi},
      year={2025},
      journal={arXiv preprint arXiv:2510.02614},
}

@misc{bansal2023universalguidancediffusionmodels,
      title={Universal Guidance for Diffusion Models}, 
      author={Arpit Bansal and Hong-Min Chu and Avi Schwarzschild and Soumyadip Sengupta and Micah Goldblum and Jonas Geiping and Tom Goldstein},
      year={2023},
      eprint={2302.07121},
      archivePrefix={arXiv},
      primaryClass={cs.CV},
      url={https://arxiv.org/abs/2302.07121}, 
}

@book{boyd2004convex,
  title     = {Convex Optimization},
  author    = {Boyd, Stephen and Vandenberghe, Lieven},
  year      = {2004},
  publisher = {Cambridge University Press},
  address   = {Cambridge, UK},
  url       = {https://web.stanford.edu/~boyd/cvxbook/}
}

@article{del2006sequential,
  title={Sequential monte carlo samplers},
  author={Del Moral, Pierre and Doucet, Arnaud and Jasra, Ajay},
  journal={Journal of the Royal Statistical Society Series B: Statistical Methodology},
  volume={68},
  number={3},
  pages={411--436},
  year={2006},
  publisher={Oxford University Press}
}

@article{jain2026diffusion,
  title={Diffusion Tree Sampling: Scalable inference-time alignment of diffusion models},
  author={Jain, Vineet and Sareen, Kusha and Pedramfar, Mohammad and Ravanbakhsh, Siamak},
  journal={Adv. Neural Inf. Process. Syst.},
  volume={38},
  pages={145576--145615},
  year={2026}
}

@inproceedings{honda2024stein,
  title={Stein variational guided model predictive path integral control: Proposal and experiments with fast maneuvering vehicles},
  author={Honda, Kohei and Akai, Naoki and Suzuki, Kosuke and Aoki, Mizuho and Hosogaya, Hirotaka and Okuda, Hiroyuki and Suzuki, Tatsuya},
  booktitle={ICRA},
  pages={7020--7026},
  year={2024},
  organization={IEEE}
}

@article{williams2017mppi,
  title={Model predictive path integral control: From theory to parallel computation},
  author={Williams, Grady and Aldrich, Andrew and Theodorou, Evangelos A},
  journal={Journal of Guidance, Control, and Dynamics},
  volume={40},
  number={2},
  pages={344--357},
  year={2017},
  publisher={American Institute of Aeronautics and Astronautics}
}

@article{Geweke1989BayesianII,
  title={Bayesian Inference in Econometric Models Using Monte Carlo Integration},
  author={John Geweke},
  journal={Econometrica},
  year={1989},
  volume={57},
  pages={1317-1339},
  url={https://api.semanticscholar.org/CorpusID:121945365}
}

@article{Boer2005,
title = "A Tutorial on the Cross-Entropy Method",
author = "\{de Boer\}, Pieter-Tjerk and Kroese, \{Dirk P.\} and Shie Mannor and Rubinstein, \{Reuven Y.\}",
year = "2005",
month = jan,
doi = "10.1007/s10479-005-5724-z",
language = "English",
volume = "134",
pages = "19--67",
journal = "Annals of operations research",
issn = "0254-5330",
publisher = "Springer",
number = "1",
}

@article{cappe2008adaptive,
  title={Adaptive importance sampling in general mixture classes},
  author={Capp{\'e}, Olivier and Douc, Randal and Guillin, Arnaud and Marin, Jean-Michel and Robert, Christian P},
  journal={Statistics and Computing},
  volume={18},
  number={4},
  pages={447--459},
  year={2008},
  publisher={Springer}
}

@misc{giaretta2026direct,
  title   = {DiRecT: Safe Diffusion-Based Planning via Receding-Horizon Denoising},
  author  = {Giaretta, Paolo and Li, Zeyang and Azizan, Navid},
  year    = {2026},
  eprint  = {2606.15359},
  archivePrefix = {arXiv},
  primaryClass = {cs.LG}
}

@article{luo2024potential,
    title={Potential based diffusion motion planning},
    author={Luo, Yunhao and Sun, Chen and Tenenbaum, Joshua B and Du, Yilun},
    journal={arXiv preprint arXiv:2407.06169},
    year={2024}
  }

@article{carvalho2025motion,
  title={Motion planning diffusion: Learning and adapting robot motion planning with diffusion models},
  author={Carvalho, Joao and Le, An Thai and Kicki, Piotr and Koert, Dorothea and Peters, Jan},
  journal={T-RO},
  volume={41},
  pages={4881--4901},
  year={2025},
  publisher={IEEE}
}

@article{zurbrugg2026dexevolve,
  title={DexEvolve: Evolutionary Optimization for Robust and Diverse Dexterous Grasp Synthesis},
  author={Zurbr{\"u}gg, Ren{\'e} and Cramariuc, Andrei and Hutter, Marco},
  journal={arXiv preprint arXiv:2602.15201},
  year={2026}
}

\end{document}